\documentclass[runningheads]{waica}
\usepackage[T1]{fontenc}
\usepackage{graphicx}
\usepackage{adjustbox}
\usepackage{subcaption}
\usepackage{xcolor}

\begin{document}
\title{Generative Transmission:
Rethinking Computation, Bandwidth, and Memory in Communication}

\titlerunning{Generative Transmission for Video Communication}


%This paper is prepared for double-blind peer review with anonymized author details

\author{{\small Xiangyu Chen, Jixiang Luo, Yuankai Fan, Haibin Huang, Chi Zhang, Xuelong Li\thanks{Corresponding author.}}}

\authorrunning{X. Chen et al.}

\institute{Institute of Artificial Intelligence (TeleAI), China Telecom
\\
\email{xuelong\_li@ieee.org}
}

\maketitle % typeset the header of the contribution

\begin{abstract}

Under the AI Flow framework, communication is shifting from transmitting fidelity-oriented information flows toward delivering task-oriented and perception-oriented token flows across heterogeneous network resources. Video communication is a fundamental component of modern information networks. However, under ultra-low-bandwidth and weak-network conditions, conventional video coding and transmission methods, which are primarily optimized for pixel-level fidelity, often struggle to balance visual usability, transmission efficiency, and robustness to unstable links. With the rapid advancement of generative models, video communication is also moving from precise signal reconstruction toward receiver-side perceptual utility and system-level usability. In this paper, we propose Generative Transmission (GenTrans) for video communication under ultra-low-bandwidth and weak-network conditions. Built upon Generative Video Compression (GVC), GenTrans formulates video transmission as a joint optimization problem involving bandwidth, computation, and memory, rather than treating it merely as a signal coding
task. By leveraging generative priors, cross-clip memory reuse, runtime state reuse, and weak-network-aware transport, GenTrans significantly reduces transmission overhead while enabling visually coherent and practically useful reconstruction. Experimental results show that GenTrans supports effective video transmission under ultra-low-bitrate and weak-network conditions, achieving improved transmission efficiency, decoding efficiency, and robustness while preserving perceptual quality.

\end{abstract}
\section{Introduction}
Video communication is a fundamental component of modern networked services and is widely used in applications such as remote collaboration, video conferencing, intelligent surveillance, unmanned systems, and edge perception. In these scenarios, visual information often needs to be transmitted continuously over bandwidth-limited and unstable links. This challenge is particularly pronounced in emergency communications, field operations, vehicular platforms, and edge networks, where video transmission may suffer from limited bandwidth, delay jitter, and random packet loss. Therefore, enabling low-overhead and robust video communication under weak-network conditions remains an important problem.

AI Flow~\cite{aiflow_edge} has recently been proposed for enabling ubiquitous intelligence over communication networks by jointly leveraging heterogeneous resources across devices, edge servers, and cloud infrastructures. Instead of viewing communication networks only as carriers of raw information, AI Flow~\cite{aiflow_edge} advocates a shift from information flow to intelligence flow, where communication is integrated with AI inference and optimized toward task-oriented utility, latency, and resource efficiency. This perspective is particularly relevant to weak-network video communication, where transmitting all pixel-level signals is often unnecessary or infeasible, while maintaining useful visual content is the primary objective.

This shift is also consistent with theories on visual information representation and task-oriented communication~\cite{yuan2025information}. From the perspective of information capacity~\cite{yuan2025information}, visual communication should not only reduce transmitted bits, but also improve the useful information carried by each unit~\cite{li2024measuring}. If the receiver can infer task-relevant visual content through computation and prior knowledge, the sender need not explicitly transmit pixel-level details. This provides a basis for computation-communication coordination and motivates generative models to trade receiver-side computation for reduced bandwidth.
Thus, Generative Transmission is not merely an engineering heuristic, but a communication paradigm grounded in theoretical foundations and physical laws governing information capacity, redundancy, and resource tradeoffs.
Similarly, positive-incentive noise~\cite{positive_incentive_noise} suggests that perturbations or approximations may be beneficial when evaluated by task-oriented utility rather than pixel-level distortion. This view encourages weak-network video communication to assess packet loss, quantization error, and generative uncertainty by receiver-side perceptual utility and task requirements.

Conventional video communication has long been dominated by fidelity-oriented coding standards such as H.264, HEVC, and VVC. These methods are primarily built upon rate-distortion optimization and have achieved remarkable success under mainstream network conditions. However, in ultra-low-bitrate and unstable-link scenarios, they often suffer from severe quality degradation, strong error propagation, and limited robustness. To overcome the performance bottleneck of predictive coding at extremely low bitrates, recent advances under the AI Flow framework\footnote{AI Flow is grounded in three laws: Law of Information Capacity, Law of Familial Model, and Law of Multi-Model Collaboration, with Generative Transmission, Familial Models, and Distributed Intelligence as representative applications.}~\cite{aiflow} have promoted research on Generative Video Compression (GVC)~\cite{gvc}. Recent advances in generative video models, represented by Sora~\cite{sora}, offer new opportunities for video compression by enabling temporally coherent video synthesis from compact latent or conditional representations. Instead of relying on pixel-accurate transmission, GVC leverages compact latent representations and strong generative priors to reconstruct visually plausible video content at the receiver, suggesting a new paradigm that improves compression efficiency by trading computation for bandwidth.

Nevertheless, existing studies on generative video compression mainly concentrate on codec-level reconstruction and have not fully addressed video communication from a unified system perspective. In particular, limited attention has been paid to the joint optimization of coding representations, transport mechanisms, and decoder states under weak-network conditions. For real-world ultra-low-bandwidth video transmission, trading computation for bandwidth alone is insufficient. Continuous video contains substantial temporal redundancy, and failing to exploit historical memory and reusable runtime states leads to unnecessary transmission and computation overhead. Meanwhile, conventional transport architectures are not designed to differentiate among generative bitstream components by role and importance, nor readily adapt transmission strategies based on network dynamics. Therefore, generative video communication should be viewed not only as a compression problem, but also as a system problem involving the joint optimization of computation, bandwidth, and memory~\cite{cbm_tradeoff}.

Motivated by this observation, we propose Generative Transmission (GenTrans), a video communication framework under ultra-low-bandwidth and weak-network conditions. Built upon Generative Video Compression (GVC), GenTrans jointly designs encoding, transport, and decoding at the system level. Specifically, at the encoder, GenTrans extracts compact visual conditioning representations that are sufficient to drive the generative model, thereby reducing the need for explicit pixel-wise transmission. At the decoder, GenTrans performs video reconstruction by combining generative priors, cross-clip memory, and reusable runtime states, reusing stable content and incrementally updating dynamic regions to reduce repeated transmission and generation cost. Furthermore, to address the inability of conventional transport architectures to perceive the internal structure and relative importance of generative bitstreams, we introduce AI Flow Transport (AFT), a transport mechanism tailored for generative video streams. AFT provides dedicated packetization and scheduling mechanisms and improves robustness against link fluctuations, queue buildup, and random packet loss through weak-network-aware transport control.

Compared with existing generative video compression schemes, GenTrans's goal is not merely to guarantee reconstruction quality and reduce video bitrate. Instead, it treats generative video transmission as an integrated AI Flow system: leveraging generative modeling to reduce explicit transmission overhead, introducing memory and runtime state reuse to reduce redundant transmission and repeated decoding, and enhancing adaptability to real-world network impairments through targeted transport mechanisms. Within this framework, the goal of video communication is no longer limited to minimizing pixel-level distortion, but extends to perceptual quality, visual intelligibility, and sustained availability under weak-network conditions. Experimental results demonstrate that GenTrans effectively reduces transmission overhead under ultra-low-bitrate and complex network conditions, while achieving better visual quality, transmission robustness, and overall usability compared to existing methods.

\section{Related Work}
Conventional video communication has been dominated by fidelity-oriented coding standards such as H.264~\cite{h264}, HEVC~\cite{hevc}, and VVC~\cite{vvc}, which primarily optimize rate-distortion performance for accurate signal reconstruction. Although highly successful in mainstream scenarios, these approaches often degrade severely under ultra-low-bitrate and unstable network conditions. To improve efficiency beyond traditional predictive coding, recent research has explored Generative Video Compression~\cite{gvc} under the AI Flow
framework~\cite{aiflow}, which reconstructs visually plausible content from compact latent representations at low bitrates. This method demonstrates the potential of trading computation for bandwidth, especially at extreme compression rates. Meanwhile, task-oriented communication has shifted attention from signal recovery to receiver-side utility, emphasizing whether transmitted content remains useful for human perception or downstream tasks. Despite these advances, previous studies typically focus either on codec-level generative reconstruction or on task-specific semantic representations, while lacking a unified system perspective for continuous video communication under weak-network conditions. In contrast, our work aims to develop a robust system that jointly optimizes computation, bandwidth, and memory for ultra-low-bandwidth video communication.

\section{Methodology}
\subsection{Motivation and Overall Framework of GenTrans}

Under the AI Flow framework~\cite{aiflow_edge,aiflow}, communication is no longer limited to the reliable delivery of raw data, but can be integrated with AI inference to transmit compact intelligence representations for task-oriented utility. Recent advances in GVC~\cite{gvc} instantiate this idea in video compression and suggest that video communication does not necessarily require the transmission of full signals at the pixel level. By sending compact latent representations and relying on the generative prior at the receiver, a system can reconstruct video content that remains semantically plausible and visually coherent at low bitrates. Compared with conventional compression pipelines, this paradigm shifts part of the reconstruction burden from transmission to generation at the receiver, thereby establishing a fundamental trade-off between computation and bandwidth.

However, under ultra-low-bandwidth and weak-network conditions, this trade-off alone does not fully address system-level redundancy. A continuous video stream is not merely a collection of isolated content units; adjacent temporal clips usually share substantial structure. Scene layout, object appearance, and background content often persist across multiple clips. If similar information is repeatedly transmitted for every clip, or if the receiver regenerates each clip from scratch, the overall communication process still contains considerable redundancy even when each individual transmission is compact. Therefore, beyond bitrate and generation cost, the reusable information accumulated over time should also be treated as a primary design factor.

Motivated by this observation, GenTrans models continuous video communication as a clip-level generative transmission process. Instead of operating on individual frames, GenTrans uses short video clips as the basic units of processing and transmission, which makes it easier to preserve local temporal structure and support cross-clip reuse. Let the input video consist of \(T\) consecutive clips:
\[
V=\{v_t\}_{t=1}^{T},
\]
where \(v_t\) denotes the \(t\)-th clip and each clip contains several consecutive frames. 
For each clip, the sender extracts a compact visual condition \(z_t\), while the receiver reconstructs the current clip by combining \(z_t\) with the memory \(m_{t-1}\) and a reusable runtime state \(s_{t-1}\):
\[
\hat{v}_t = G(z_t, m_{t-1}, s_{t-1}),
\]
where \(G(\cdot)\) denotes the conditional generation process at the receiver. Here, \(z_t\) provides the incremental information needed for the current clip, \(m_{t-1}\) stores reusable visual content accumulated from previous clips, and \(s_{t-1}\) denotes a compatible prior runtime state available before decoding the current clip.

Under this formulation, the objective is no longer limited to minimizing the transmission cost of each clip in isolation. Instead, GenTrans jointly considers the costs of transmission, generation, and state maintenance. This can be abstractly written as:
\[
\min \sum_{t=1}^{T}\left(
\lambda_b B_t + \lambda_c C_t + \lambda_m M_t + \lambda_u U_t
\right),
\]
where \(B_t\) is the transmission cost for clip \(v_t\), \(C_t\) is the computation cost for reconstructing it at the receiver, \(M_t\) denotes the cost of maintaining memory and runtime states, and \(U_t\) measures the utility loss of the reconstructed result. The utility term is not restricted to pixel distortion; it may also reflect degraded perceptual quality, weakened temporal consistency, or reduced overall usability of the reconstructed video. The coefficients \(\lambda_b\), \(\lambda_c\), \(\lambda_m\), and \(\lambda_u\) balance the relative importance of these terms.

Overall, GenTrans is not designed to perform stronger per-clip generation in isolation. Its goal is to make continuous video transmission more efficient by jointly using current conditions, historical memory, and runtime states, so as to reduce repeated transmission and repeated computation while maintaining stable visual output under challenging network conditions.

\subsection{Reducing Bandwidth via Generative Memory}

Continuous video often contains a large amount of stable content that can be reused across clips. In fixed-view or low-motion scenarios, for example, background regions, object appearance, and overall spatial layout usually remain similar over short periods of time. This means that not all visual information needs to be retransmitted for the current clip. Part of it has already been established during the reconstruction of previous clips and can continue to support subsequent generation.

To exploit this property, GenTrans introduces Generative Memory (GenMem), a cross-clip memory mechanism that preserves visual information likely to remain useful in future reconstruction. In GenTrans, this memory is maintained across continuous transmission so that both the sender and the receiver can operate with a shared memory context over time. As a result, the current clip no longer depends solely on newly received conditions; it can also be partially supported by the accumulated memory. From the sender's perspective, this reduces the need to repeatedly describe already stabilized content, allowing transmission to focus on indicating which memory content should be reused, together with changing regions, newly appeared content, and information that cannot be reliably recovered from memory alone.

Let \(h_t\) denote a high-level visual representation extracted from the current clip \(v_t\). We approximate it as the combination of a reusable historical component and an incremental component:
\[
h_t \approx R(m_{t-1}, \pi_t) + \Delta_t,
\]
where \(R(m_{t-1}, \pi_t)\) denotes the reference representation retrieved from the memory \(m_{t-1}\) under a matching relation \(\pi_t\), and \(\Delta_t\) captures the newly introduced or changed information in the current clip. Here, \(\pi_t\) can be understood as a compact memory reference that specifies how the current clip relates to the previously accumulated memory, while \(\Delta_t\) corresponds to the incremental information that still needs to be conveyed. This expression is not intended as a strict signal decomposition. Rather, it describes the communication principle behind GenTrans: if a substantial part of the current content can already be supported by memory, then the information that truly needs to be transmitted should mainly correspond to the memory reference and the remaining increment.

After each clip is reconstructed, the memory is updated so that it can continue to serve later clips. This process is written as:
\[
m_t = \Phi(m_{t-1}, \hat{v}_t, z_t),
\]
where \(\Phi(\cdot)\) denotes the memory update function. Based on the reconstructed clip and the received condition, it retains, supplements, or replaces memory entries to maintain a temporally useful pool of visual priors. We do not impose a specific memory structure here; what matters is its role in reducing repeated transmission and providing stable support for future generation.

This mechanism is fundamentally different from reference-frame prediction in conventional video coding. Traditional predictive coding mainly relies on signal-domain prediction and residual compensation, whereas the memory in GenTrans is intended to support generation through reusable high-level visual representations. In other words, what is reused here is not an exact signal reference, but compact prior information that remains helpful for reconstructing later clips. This makes the mechanism more compatible with generative receivers and more suitable for operation at extremely low bitrates.

\subsection{Trading Memory for Computation}

Reducing bitrate alone does not automatically make a generative transmission system efficient as a whole. At the receiver, if every newly arrived clip still requires a full generation process from scratch, continuous communication remains computationally expensive and may introduce substantial latency. This issue becomes more pronounced when the current decoding task shares significant visual structure or generation context with previously reconstructed content, since rebuilding all internal representations from an empty state leads to repeated inference over highly similar information.

To alleviate this overhead, GenTrans introduces \emph{state reuse} at the receiver. The key idea is to preserve reusable runtime states produced during previous decoding processes, and to use a compatible prior state to initialize the current reconstruction whenever possible. In this way, the receiver does not always start from an empty state. Instead, it continues reconstruction from a partially established generative context. The purpose is not merely to cache previous outputs, but to reuse intermediate context, internal representations, or high-confidence generation results that remain valuable for subsequent reconstruction.

Let \(s_{t-1}\) denote a compatible prior runtime state available before decoding the current clip. The generation of the current clip can then be written as:
\[
\hat{v}_t = G(z_t, m_{t-1}, s_{t-1}),
\]
and the corresponding state update is:
\[
s_t = \Psi(s_{t-1}, z_t, \hat{v}_t),
\]
where \(\Psi(\cdot)\) denotes the state transition function. 

Under this formulation, the receiver is not solving a sequence of isolated generation problems. Instead, it performs a continuous or state-connected reconstruction process in which useful generative context can be carried across time or reused across sufficiently similar decoding instances.

From a resource perspective, this mechanism effectively trades limited storage and state maintenance for lower repeated computation. Let \(C_t^{\mathrm{full}}\) denote the computation cost of reconstructing the current clip from scratch, and let \(C_t^{\mathrm{reuse}}\) denote the actual cost when state reuse is enabled. In general, we have:
\[
C_t^{\mathrm{reuse}} \leq C_t^{\mathrm{full}}.
\]
When the prior state is highly compatible with the current condition, the reduction can be substantial. In other words, by preserving compact but useful runtime states, the receiver can avoid repeatedly performing full inference on already established generative context, thereby reducing latency and improving the decoding efficiency.

\subsection{Robust Transmission under Weak-Network Conditions}

Ultra-low-bitrate video communication is typically accompanied by weak-network conditions such as bandwidth fluctuations, random packet loss, and delay jitter. In GenTrans, the receiver does not obtain a complete signal for pixel-accurate inversion, but rather compact visual conditions that drive generative reconstruction. Therefore, system usability under weak-network conditions depends not only on whether the network can deliver all data intact, but also on whether the model can maintain perceptually acceptable outputs when conditions are incomplete. Based on this observation, the weak-network design of GenTrans consists of two complementary aspects: packet-loss-robust training on the model side and transmission support for generative visual flows on the network side.

First, on the model side, GenTrans explicitly simulates condition missing during training in order to improve the model's tolerance to link perturbations. Let the generative conditions corresponding to the current clip be
\[
z_t=\{z_t^{(1)}, z_t^{(2)}, \dots, z_t^{(L)}\},
\]
where different components denote condition tokens of different types or granularities. During training, the system randomly drops a subset of these components and retains only a subset \(\tilde{z}_t \subseteq z_t\) as input, while requiring the receiver to reconstruct the current clip under incomplete conditions:
\[
\hat{v}_t = G(\tilde{z}_t, m_{t-1}, s_{t-1}).
\]
In essence, this strategy injects random packet loss into the training process, enabling the generator to compensate for missing information by leveraging historical memory, runtime states, and the remaining conditions, thereby achieving more stable reconstruction in actual weak-network conditions.

Second, on the network side, GenTrans introduces a dedicated support protocol for generative video transmission, referred to as AI Flow Transport (AFT). Rather than following conventional uniform transport of video bitstreams, AFT is designed around the organization and delivery of generative condition flows. Its core objective is to apply differentiated handling to condition tokens with different levels of importance under unstable links, so as to strike a better balance between reliability and latency from a perceptual perspective. Specifically, AFT adapts the generative bitstream structure, congestion control, packet retransmission, and forward redundancy mechanisms at the system level, allowing critical conditions to receive prioritized protection instead of treating all transmitted contents as equivalent data. The model-side robust training improves the system's ability to continue generating under missing conditions, while AFT further improves the likelihood that critical conditions arrive in time; together, they form the foundation of robust transmission in real weak-network conditions.

\subsection{Summary and Discussion}

\begin{figure}[t]
    \centering
    \includegraphics[width=0.98\linewidth]{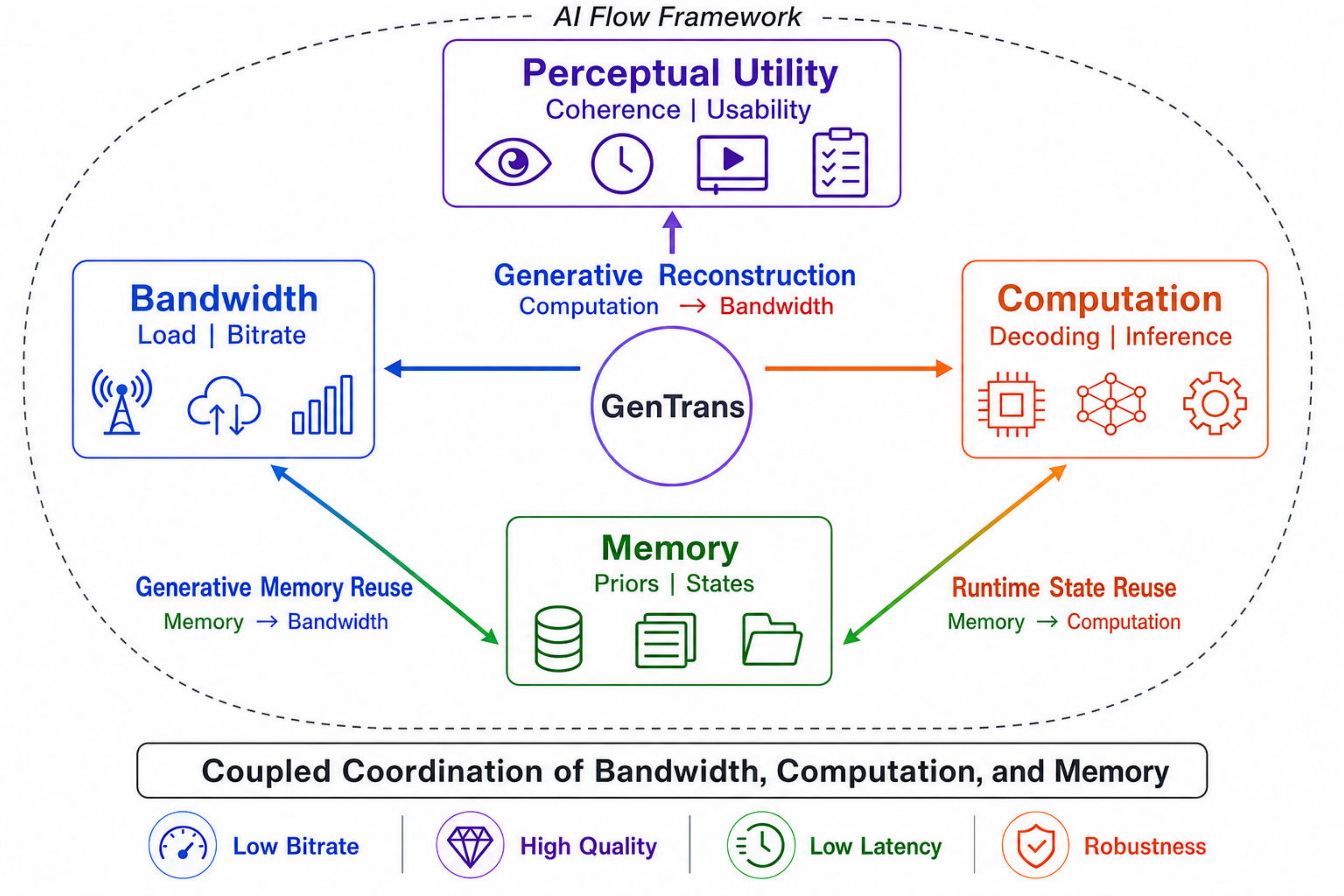}
    \caption{Overview of GenTrans as a coupled coordination framework of bandwidth,
    computation, and memory under the AI Flow framework.}
    \vspace{-2pt}
    \label{fig:GenTrans_framework}
\end{figure}

Overall, the key contribution of GenTrans is not an isolated coding or generation module, but a system-level reorganization of the key resources in generative
video communication under the AI Flow framework. As illustrated in Fig.~\ref{fig:GenTrans_framework}, GenTrans treats bandwidth, computation, and memory as coupled resources, and coordinates them toward perceptual utility. 

More specifically, GenTrans consists of three mutually supportive mechanisms. First, generative reconstruction enables the system to replace traditional high-redundancy
signal transmission with a small number of compact visual conditions, corresponding to “trading computation for bandwidth.” Second, cross-clip memory reuse prevents stable content from being repeatedly described during continuous communication, corresponding to “using memory to reduce transmission.” Third, runtime state reuse allows the receiver to avoid performing fully independent
inference for every clip. This reuse arises most naturally between temporally adjacent clips, but can also be extended to historically matched decoding contexts when compatible cached states are available, corresponding to “trading memory for computation.” 

This resource coordination also changes the optimization objective of video communication. Conventional video coding is mainly driven by rate-distortion optimization and pixel-level fidelity. In contrast, GenTrans focuses on maintaining perceptual usability under practical constraints. Stronger generation can reduce transmission load but increases receiver-side computation; more aggressive memory reuse can lower bitrate but requires reliable memory management; and state reuse can reduce latency but depends on the compatibility of cached
runtime states. Therefore, GenTrans does not optimize any single resource in isolation, but seeks a balanced operating point among bandwidth, computation, and memory. Besides, model-side robust training and AFT-based transmis-
sion jointly ensure that these generative mechanisms can continue to function over unstable real-world links, so that the system is effective not only under ideal conditions but also remains usable in practical weak-network deployment.

In this sense, GenTrans should be regarded as a flexible system framework rather than a fixed codec configuration. In extremely bandwidth-constrained scenarios, the system can rely more on generative reconstruction and perceptually
prioritized AFT delivery. In slowly changing scenes, cross-clip memory can provide larger transmission savings. In latency-sensitive edge scenarios, runtime state reuse can reduce repeated decoding overhead. Under unstable links, robust training and AFT allow the system to degrade gracefully rather than collapse abruptly. This makes GenTrans suitable for continuous video communication under diverse weak-network and resource-constrained conditions.

\section{Results}
We evaluate GenTrans from three dimensions directly corresponding to the system design goals: continuous transmission bandwidth, receiver decoding efficiency, and robustness in weak-network conditions. First, we examine whether GenMem can reduce redundant transmissions in continuous video communication, thereby further reducing the overall transmission bitrate. Second, we analyze whether runtime state reuse can reduce the overhead of repeated generation at the receiver, thereby improving decoding efficiency in continuous transmission. Finally, we evaluate how the proposed weak-network robustness design affects system stability and availability under random packet loss conditions.

\subsection{Bandwidth Reduction with GenMem}

We first evaluate the bandwidth efficiency of GenTrans, with a particular focus on whether it can substantially reduce transmission cost while maintaining perceptual reconstruction quality under ultra-low-bitrate conditions. Experiments are conducted on the MCL-JCV~\cite{mcl_jcv} 720p dataset. We use bits per pixel (bpp) to measure transmission bitrate and LPIPS~\cite{lpips} to evaluate perceptual reconstruction quality. We compare GenTrans with HEVC~\cite{hevc} and with GVC~\cite{gvc}, a prior generative compression method that does not use GenMem, to isolate the contribution of GenMem.

\begin{table}[t]
    \centering
    \caption{Quantitative comparison on the MCL-JCV 720p dataset.}
    \renewcommand{\arraystretch}{1.2}
    \setlength{\tabcolsep}{20pt}
    \begin{tabular}{l c c}
        \hline
        Method & Bitrate (bpp) & LPIPS ($\downarrow$) \\ \hline
        GVC~\cite{gvc} & 0.008 & 0.180 \\
        HEVC~\cite{hevc} & 0.008 & 0.278 \\
        \textbf{GenTrans (with GenMem)} & 0.006 & 0.185 \\
        HEVC~\cite{hevc} & 0.006 & 0.328 \\ \hline
    \end{tabular}
    \label{tab:bandwidth_main_results}
\end{table}

Table~\ref{tab:bandwidth_main_results} reports representative quantitative results at low-bitrate operating points. Under comparable bitrate budgets, GenTrans consistently achieves substantially better perceptual quality than HEVC. For example, at approximately 0.006~bpp, GenTrans attains an LPIPS of 0.185, which is markedly lower than that of HEVC at the same bitrate. This result indicates that, under highly constrained transmission budgets, GenTrans preserves perceptual fidelity much more effectively than conventional signal-level compression. More importantly, comparison with the GVC baseline demonstrates the additional benefit brought by GenMem. Relative to GVC, which operates at 0.008~bpp with an LPIPS of 0.180, GenTrans achieves a comparable perceptual quality (0.185 LPIPS) at only 0.006~bpp. This corresponds to an additional bitrate reduction of approximately 25\%, with only a marginal change in perceptual quality. These results suggest that GenMem can effectively reduce redundant transmission by reusing previously accumulated visual memory across temporally related video clips.

\begin{figure}[h]
    \centering
    \includegraphics[width=0.9\linewidth]{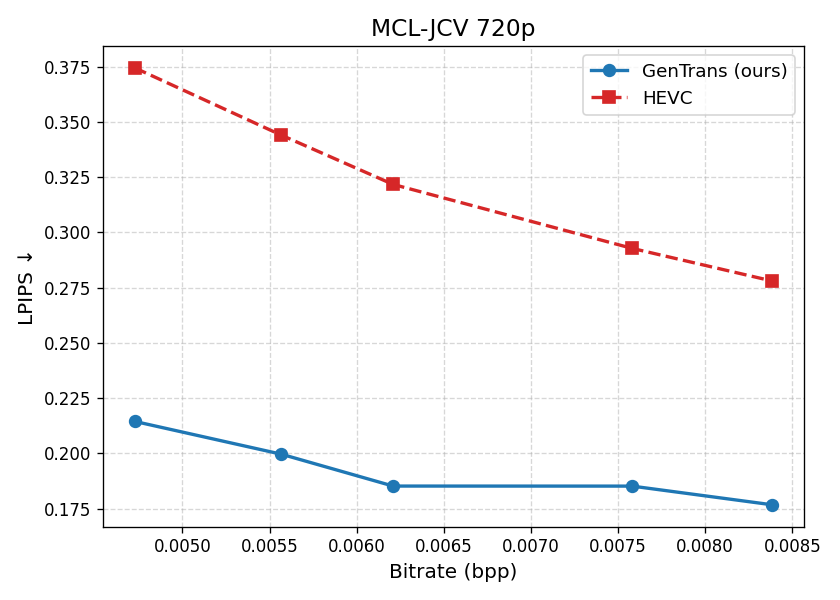}
    \caption{Bitrate-LPIPS curves on MCL-JCV 720p dataset.}
    \label{fig:rp_curve}
\end{figure}

To further compare GenTrans and HEVC over a wider range of operating points, we plot bitrate-LPIPS curves in Figure~\ref{fig:rp_curve}. GenTrans consistently achieves lower LPIPS than HEVC at the same bitrate across the ultra-low-bitrate regime, confirming its clear advantage in rate-perception trade-off.

The results above reflect average performance over the benchmark dataset. In practice, the benefit of GenMem becomes more pronounced in scenes with stable structure, limited camera motion, or low temporal variation, where background content and appearance patterns can be retained and reused across video clips. As a representative example, Table~\ref{tab:bandwidth_case_study} presents results on \texttt{VideoSRC30} from MCL-JCV. In this case, GenTrans reduces the bitrate from 0.0103~bpp to 0.0035~bpp while maintaining similar perceptual quality to GVC, with LPIPS values of 0.0855 and 0.0866, respectively. This corresponds to a bitrate reduction of about 66\%, indicating that GenMem can substantially reduce retransmission overhead when scene content remains stable over time.

\begin{table}[ht]
    \centering
    \caption{Case study in a low-motion scenario.}
    \renewcommand{\arraystretch}{1.2}
    \setlength{\tabcolsep}{20pt}
    \begin{tabular}{l c c}
        \hline
        Method & Bitrate (bpp) & LPIPS ($\downarrow$) \\ \hline
        GVC~\cite{gvc} & 0.0103 & 0.0855 \\
        \textbf{GenTrans (with GenMem)} & 0.0035 & 0.0866 \\ 
        \hline
    \end{tabular}
    \label{tab:bandwidth_case_study}
\end{table}

\begin{figure*}[t]
    \centering
    % Row 1
    \begin{subfigure}[b]{0.32\textwidth}
        \centering
        \includegraphics[width=\linewidth]{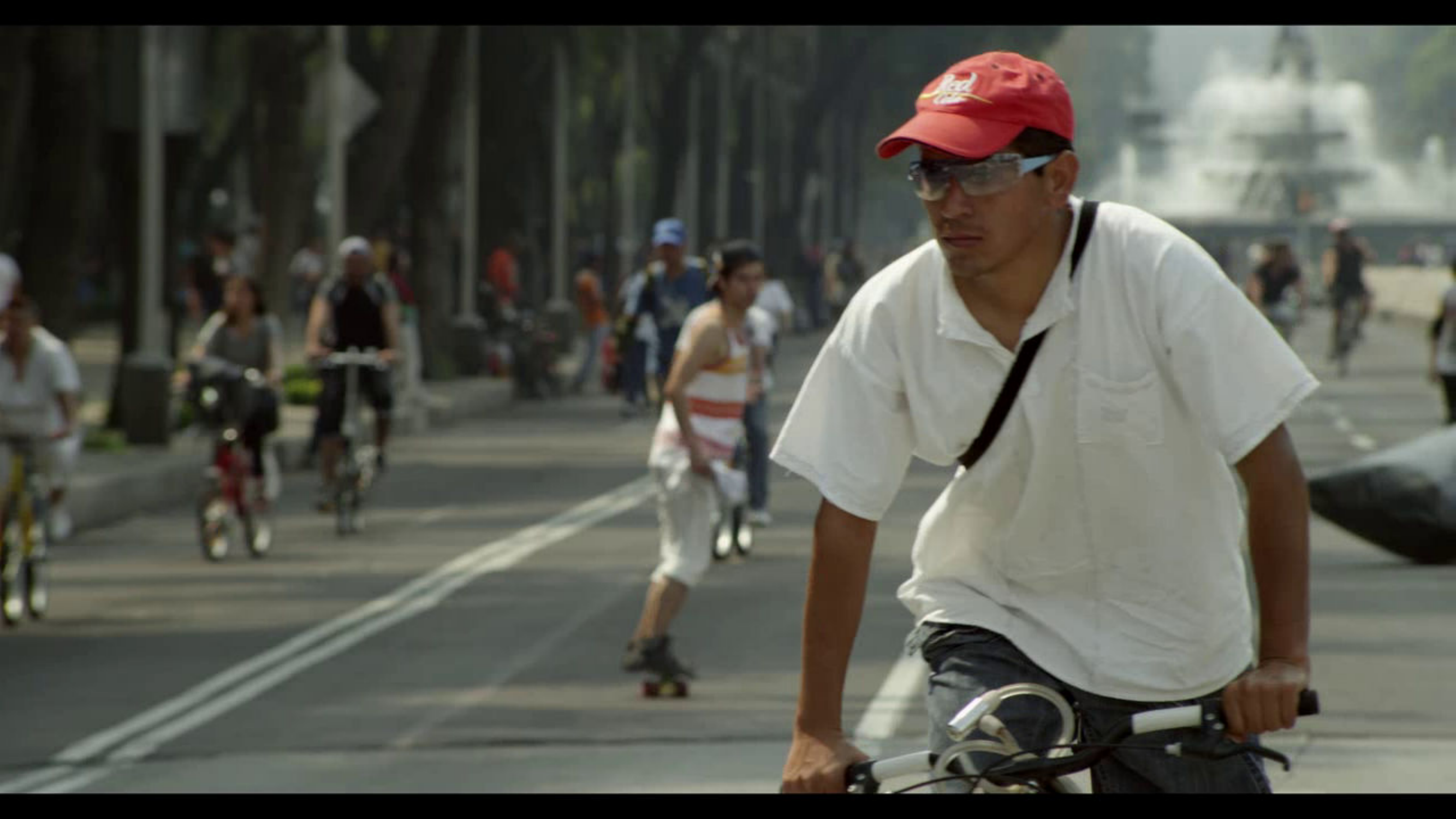}
        \caption*{VideoSRC02 - bpp: 0.0046}
    \end{subfigure}
    \hfill
    \begin{subfigure}[b]{0.32\textwidth}
        \centering
        \includegraphics[width=\linewidth]{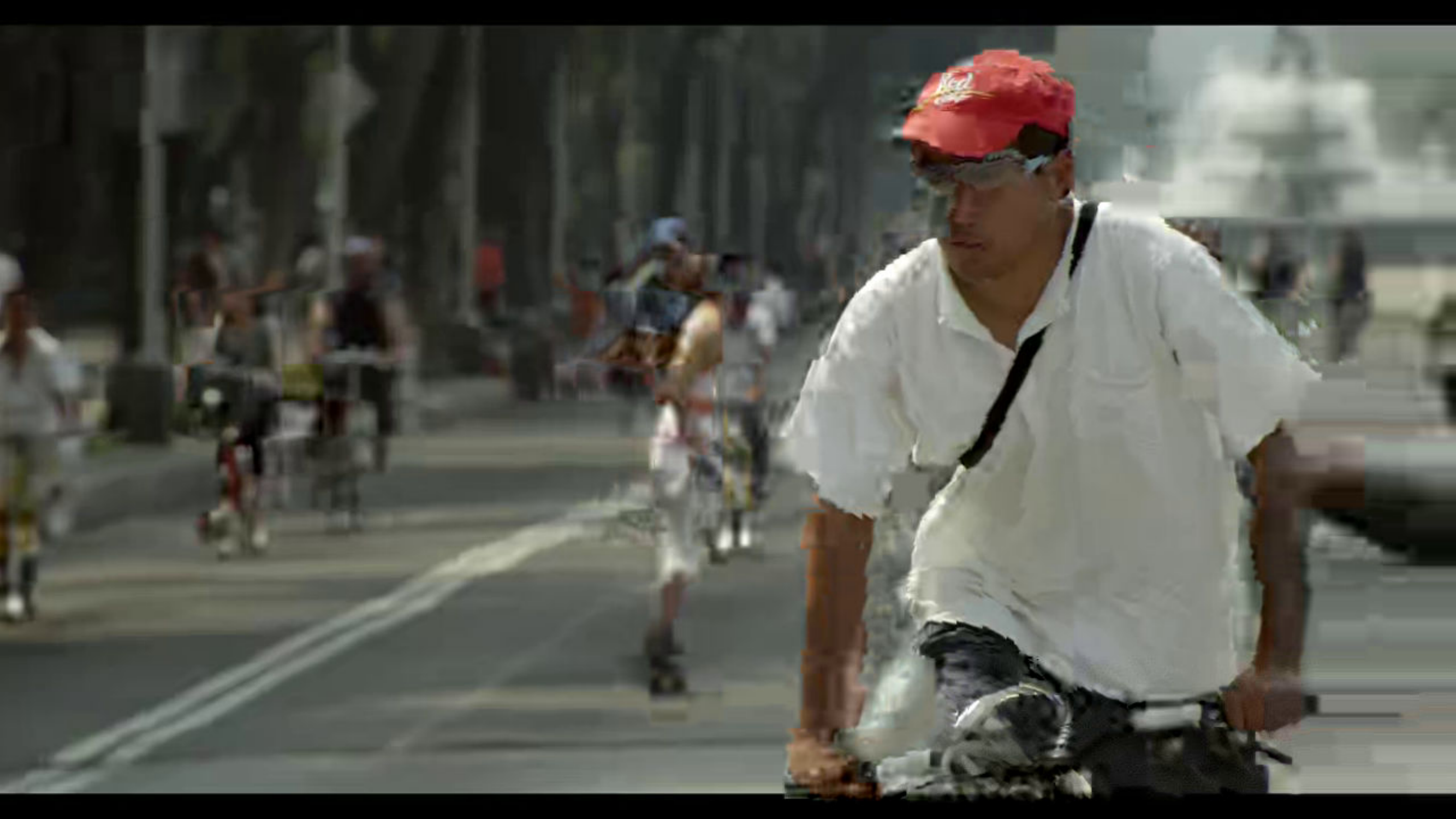}
        \caption*{HEVC - LPIPS: 0.3804}
    \end{subfigure}
    \hfill
    \begin{subfigure}[b]{0.32\textwidth}
        \centering
        \includegraphics[width=\linewidth]{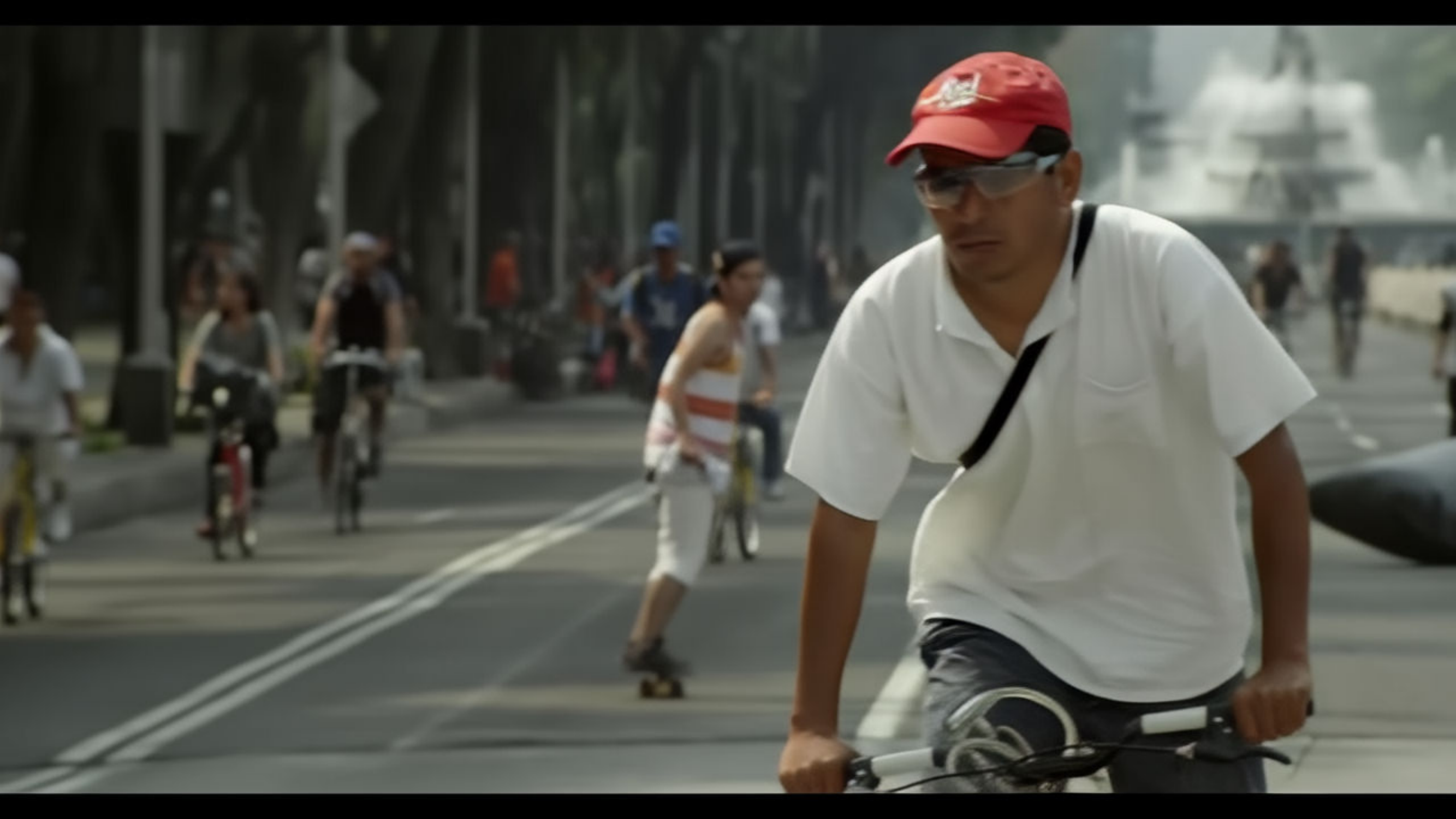}
        \caption*{GenTrans - LPIPS: 0.1718}
    \end{subfigure}
    \vspace{0.5em}
    % Row 2
    \begin{subfigure}[b]{0.32\textwidth}
        \centering
        \includegraphics[width=\linewidth]{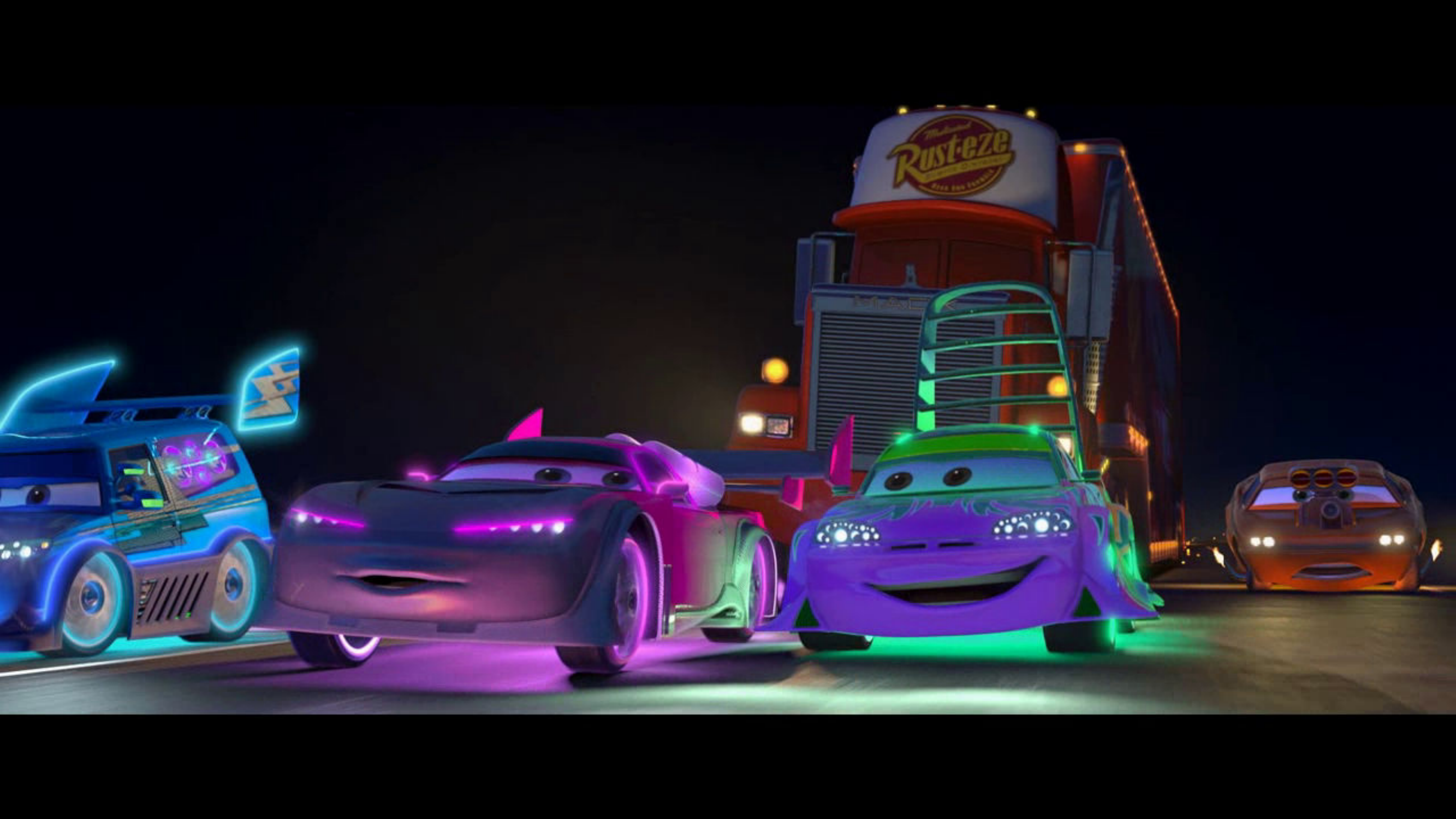}
        \caption*{VideoSRC24 - bpp: 0.0055}
    \end{subfigure}
    \hfill
    \begin{subfigure}[b]{0.32\textwidth}
        \centering
        \includegraphics[width=\linewidth]{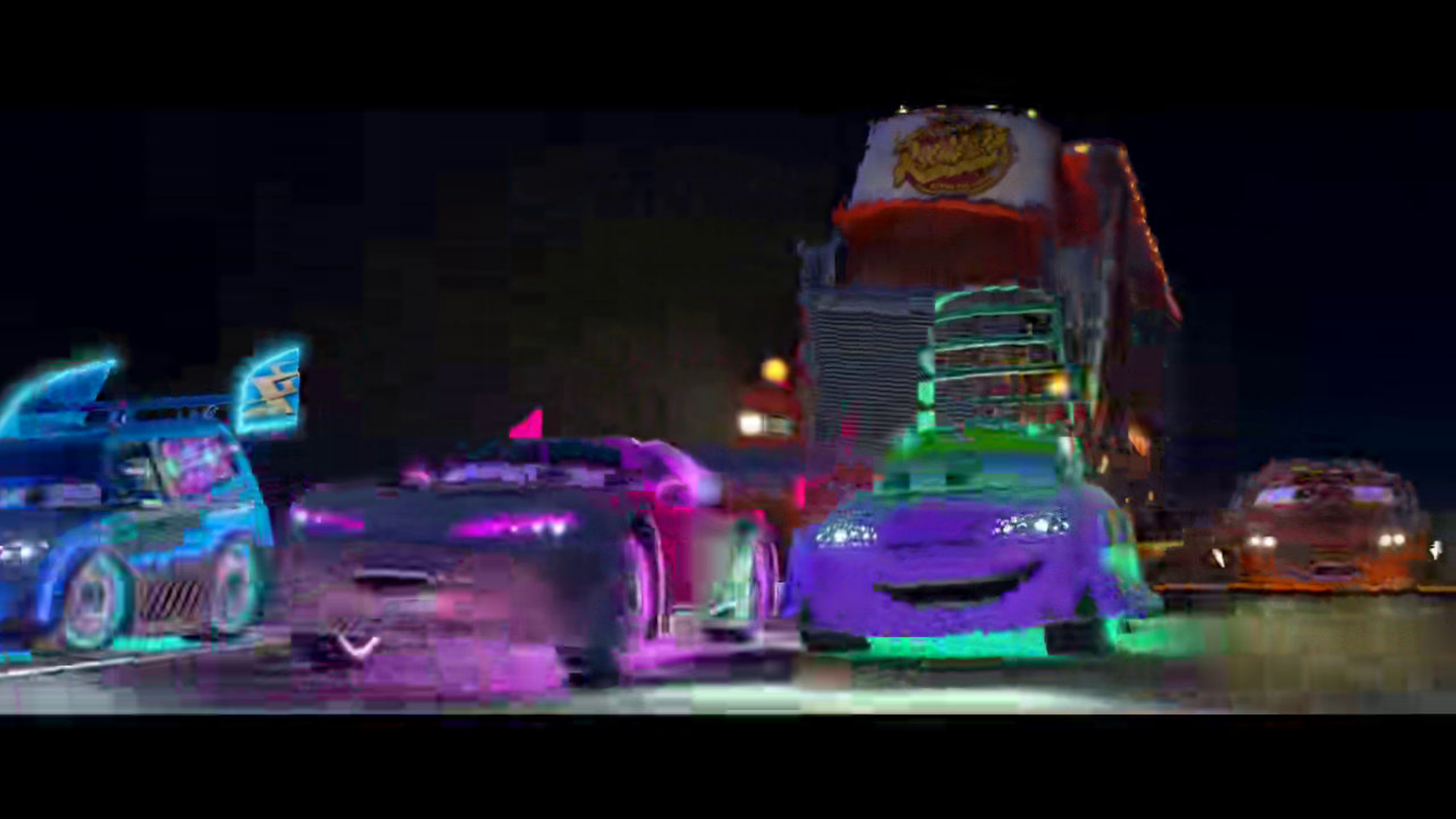}
        \caption*{HEVC - LPIPS: 0.1682}
    \end{subfigure}
    \hfill
    \begin{subfigure}[b]{0.32\textwidth}
        \centering
        \includegraphics[width=\linewidth]{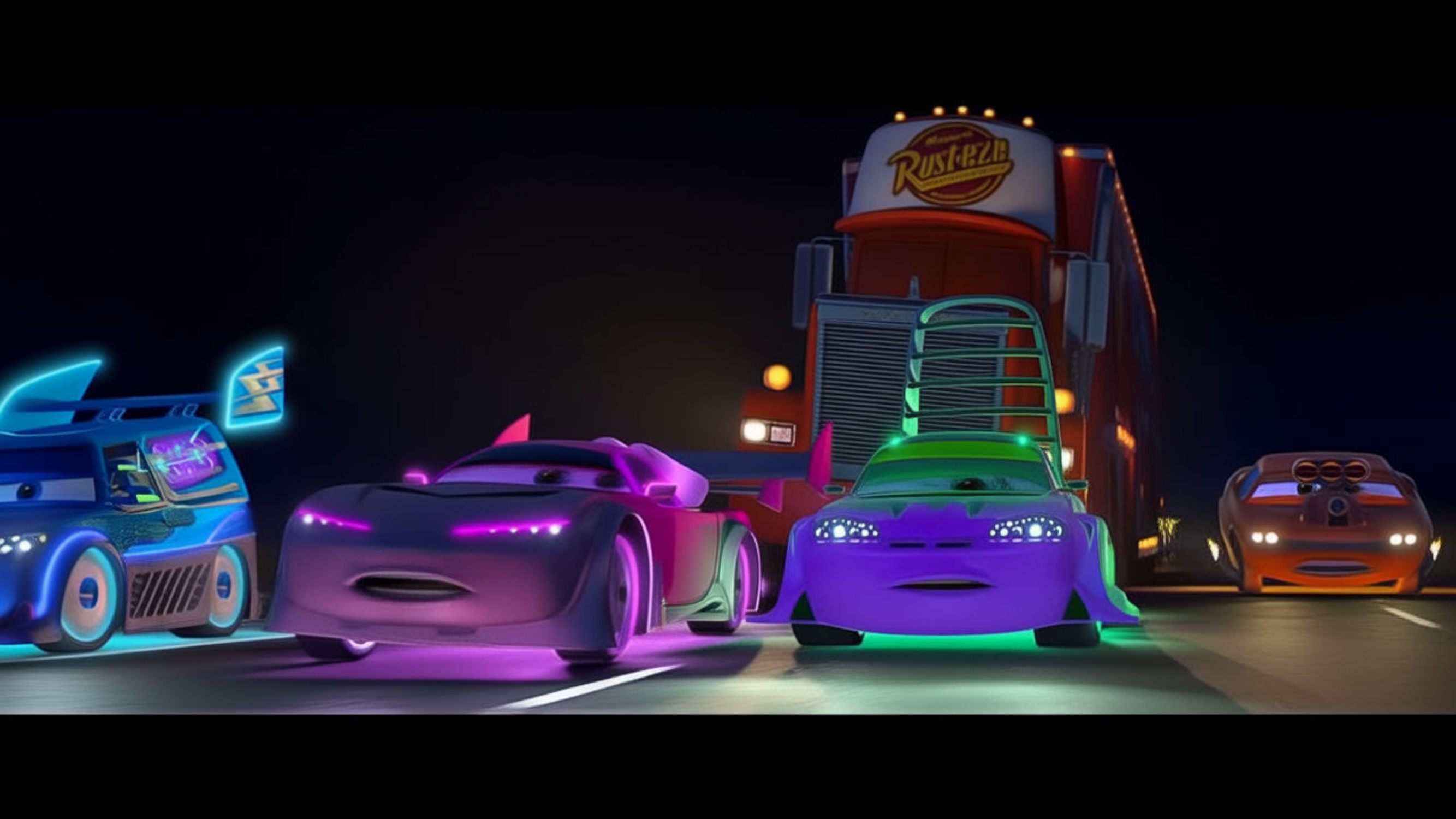}
        \caption*{GenTrans - LPIPS: 0.0836}
    \end{subfigure}
    \caption{Visual comparison between HEVC and GenTrans at similar bitrates.}
    \label{fig:bandwidth_visual}
\end{figure*}

Figure~\ref{fig:bandwidth_visual} further provides visual comparisons at similar bitrate levels. In both examples, GenTrans preserves scene structure, visual naturalness, and semantically important content much better than HEVC. In contrast, HEVC exhibits severe compression artifacts, structural blurring, and loss of fine details under ultra-low-bitrate conditions. These visual observations are consistent with the quantitative trends in Table~\ref{tab:bandwidth_main_results} and Figure~\ref{fig:rp_curve}, further confirming that GenTrans significantly outperforms conventional video coding under extreme bitrate constraints.

\subsection{Decoding Efficiency via State Reuse}

Reducing transmission bandwidth alone does not guarantee system-level efficiency in generative video communication. At the receiver, if each decoding instance still starts from scratch and performs full inference from an empty state, the computational cost and response latency remain high. To address this issue, we further evaluate whether state reuse can reduce redundant computation and improve overall decoding efficiency.

We conduct the evaluation on the dataset of real-world decoding instances collected from TeleStudio\footnote{\url{https://telestudio.teleagi.cn/generatevideo/home}}, and randomly sample 100 instances for analysis. Table~\ref{tab:state_reuse_results} summarizes the results. Under a representative GenTrans setting, state reuse reduces the average decoding latency from 62.4~s to 41.2~s, achieving an approximately \(1.5\times\) speedup. Meanwhile, the reuse rate reaches 37\%, indicating that compatible prior states occur frequently enough in realistic workloads to produce significant system-level gains. These results suggest that a compatible prior runtime state can serve as an effective initialization for the current decoding instance, thereby avoiding unnecessary repeated inference. Therefore, this design of GenTrans can effectively mitigate a key concern of generative communication, namely that reduced transmission bandwidth may come at the cost of excessive inference overhead.

\begin{table}[t]
    \centering
    \caption{Decoding efficiency with state reuse under a representative setting.}
    \begin{adjustbox}{width=.8\linewidth}
        \begin{tabular}{l c@{\hspace{10pt}} c@{\hspace{10pt}} c}
            \hline
            Method & Latency (s) & Speedup & Reuse Rate \\ \hline
            Baseline (w/o reuse) & 62.4 & -- & -- \\
            State reuse & 41.2 & 1.5$\times$ & 37\% \\ \hline
        \end{tabular}
    \end{adjustbox}
    \label{tab:state_reuse_results}
\end{table}

\subsection{Robustness under Weak-Network Conditions}

\begin{table}[t]
  \centering
  \caption{Reconstruction quality under different packet loss ratios.}
  \label{tab:lpips_packet_loss}
  \normalsize
  \renewcommand{\arraystretch}{1.5}
  \setlength{\tabcolsep}{13pt}
  \begin{tabular}{l|c|c|c|c}
    \hline
    Drop Ratio & 0\% & 10\% & 30\% & 50\% \\ \hline
    LPIPS ($\downarrow$) & 0.1748 & 0.1821 & 0.2030 & 0.3296 \\ \hline
  \end{tabular}
\end{table}

\begin{figure}[t]
\centering
% 左边：(a)
\begin{minipage}[c]{0.42\textwidth}
    \centering
    \includegraphics[width=\textwidth]{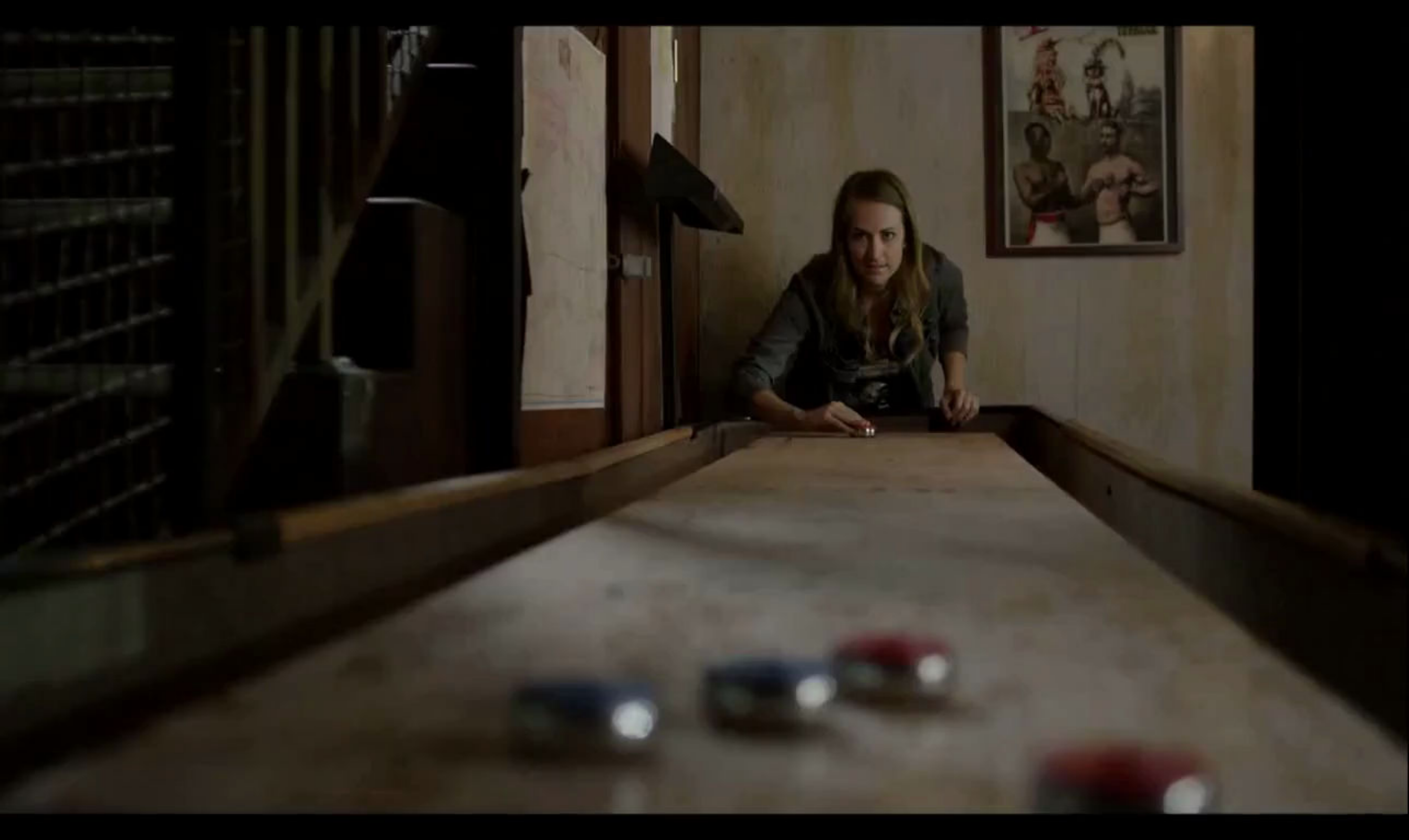}\\[2pt]
    \small (a) Original Frame
\end{minipage}
\hfill
% 右边：2x2
\begin{minipage}[c]{0.54\textwidth}
    \centering
    \begin{tabular}{cc}
        \includegraphics[width=0.47\textwidth]{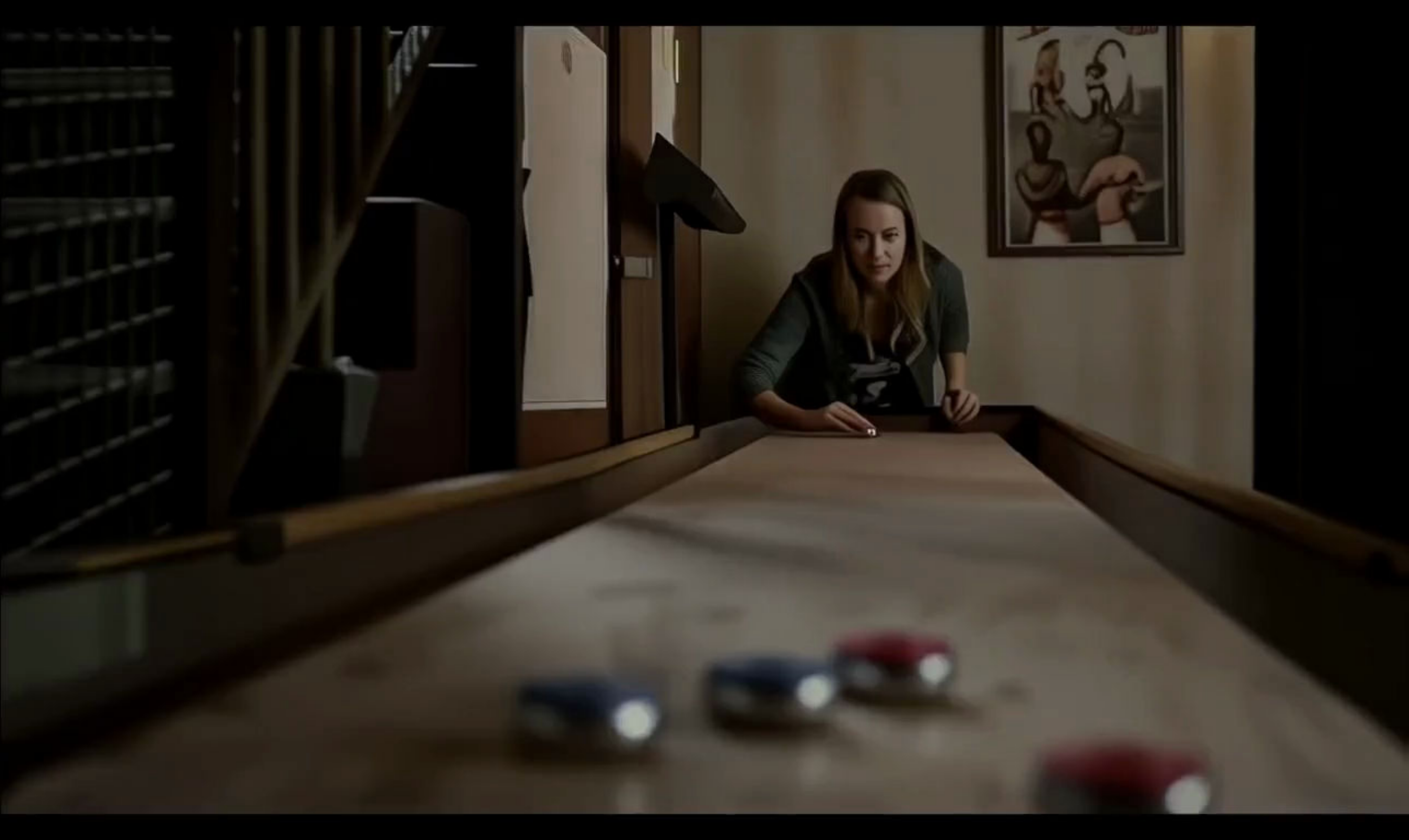} &
        \includegraphics[width=0.47\textwidth]{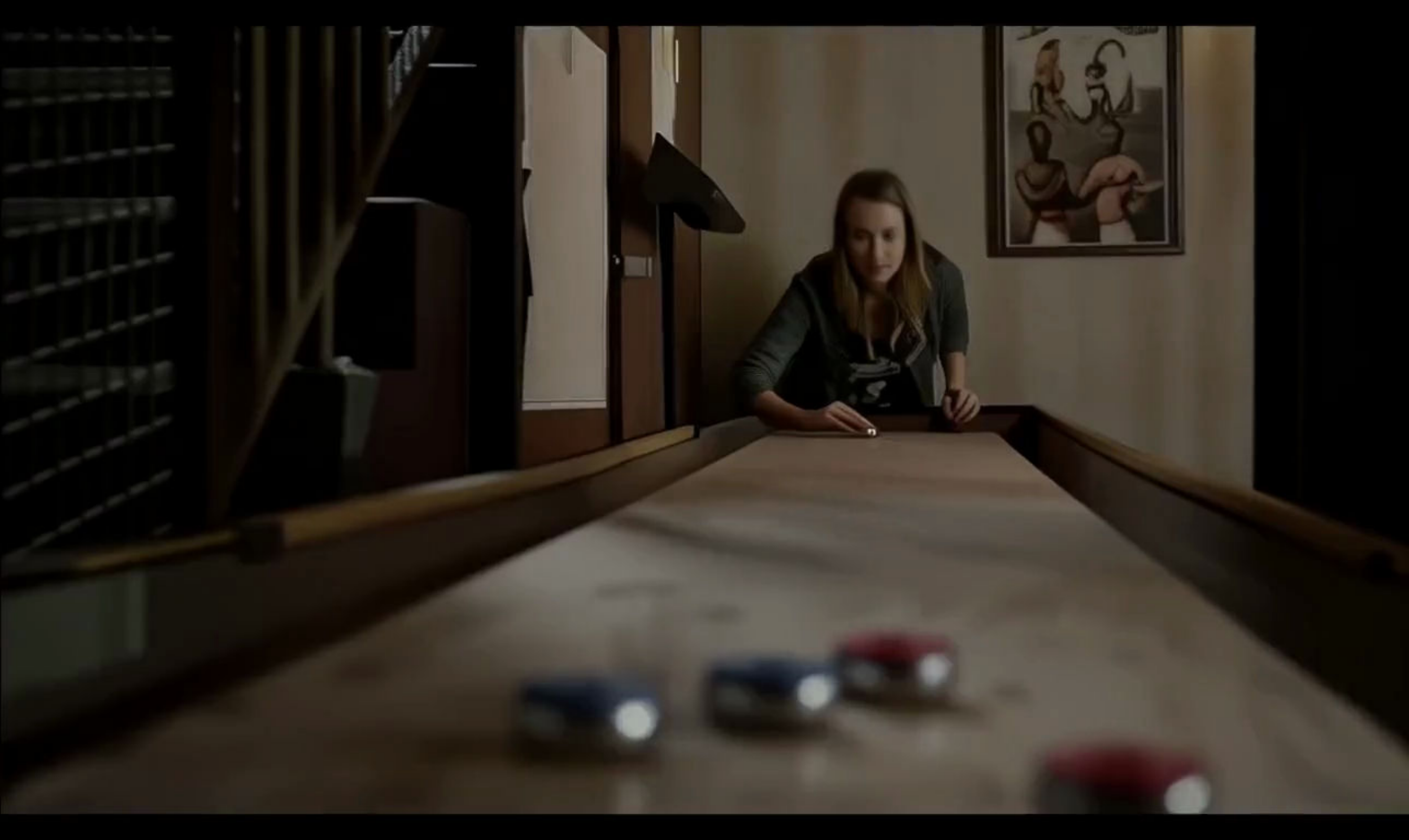} \\[2pt]
        \small (b) 0\% Loss & \small (c) 10\% Loss \\[6pt]

        \includegraphics[width=0.47\textwidth]{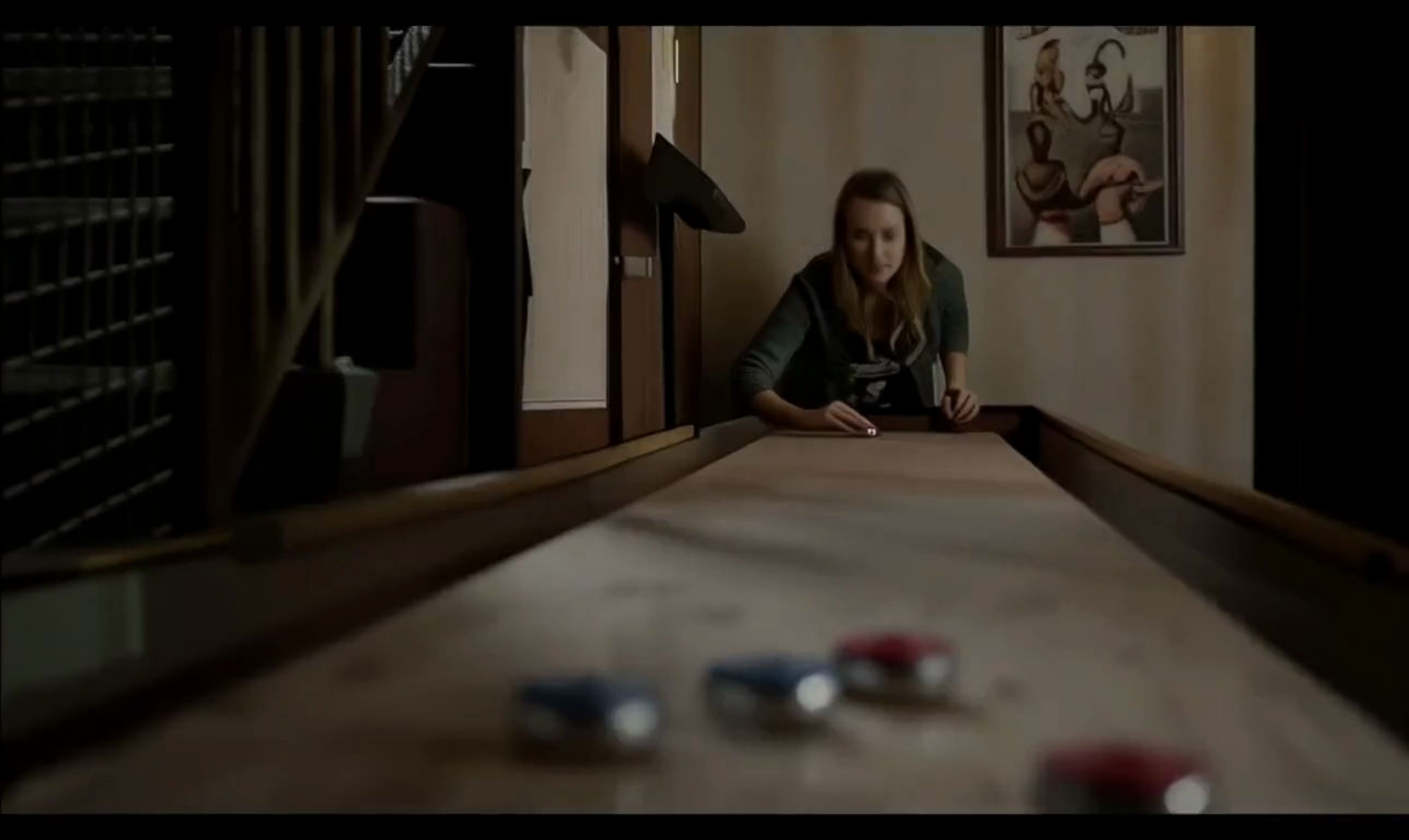} &
        \includegraphics[width=0.47\textwidth]{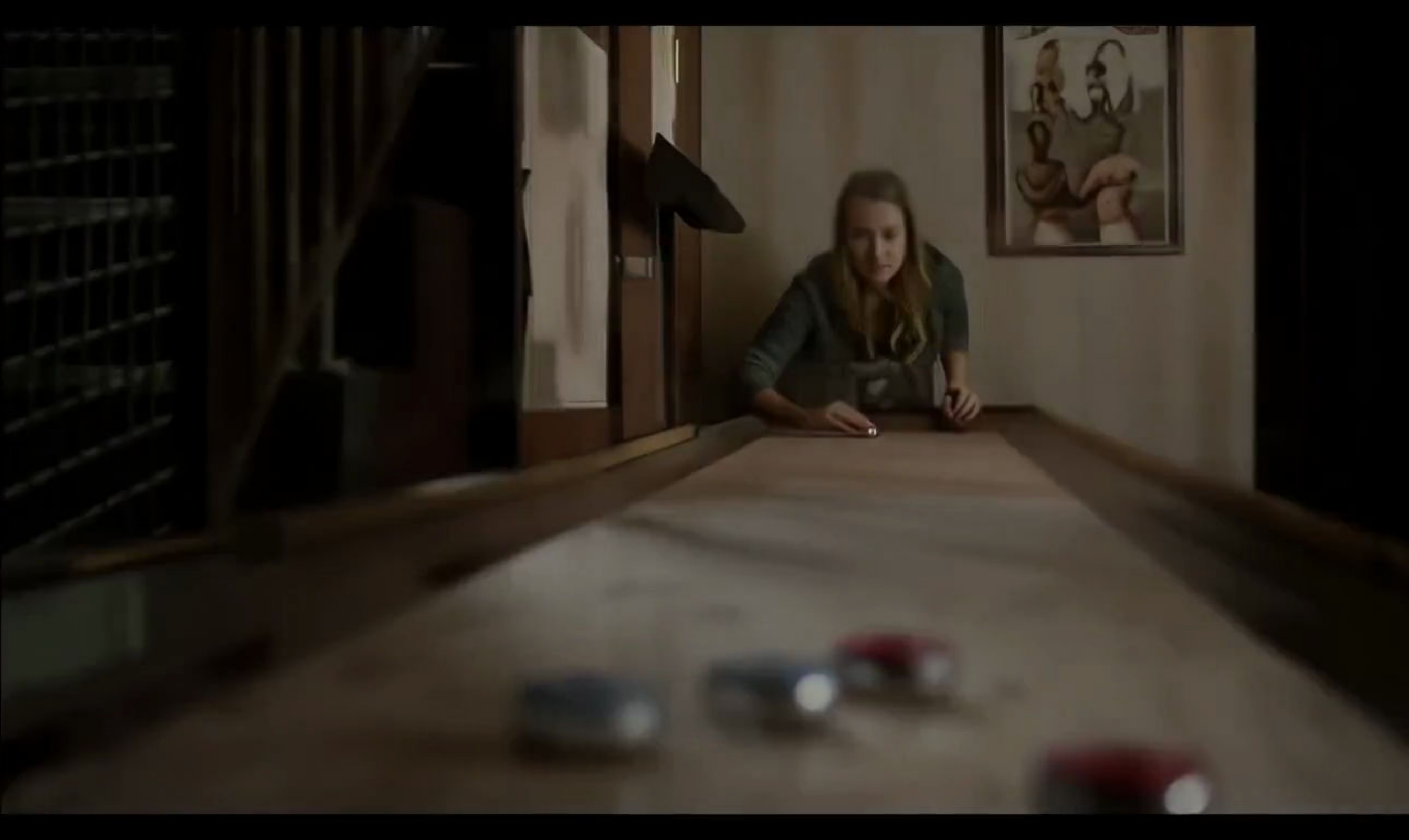} \\[2pt]
        \small (d) 30\% Loss & \small (e) 50\% Loss
    \end{tabular}
\end{minipage}
\caption{Reconstruction results under different packet loss rates.}
\vspace{-10pt}
\label{fig:packet_loss_reconstruction}
\end{figure}

In addition to bandwidth and decoding efficiency, practical deployment of GenTrans also depends on whether it can remain usable under unstable network conditions. Since generative video transmission relies on conditional information at the receiver side, packet loss does not merely introduce signal distortion; it may directly damage the completeness of the received conditions and thus affect the generation process itself. We therefore further evaluate the robustness of the proposed framework under lossy network conditions, focusing on how reconstruction quality changes as packet loss increases.

We conduct experiments on the MCL-JCV dataset at a fixed bitrate of about 0.01 bpp, and evaluate perceptual reconstruction quality under different packet loss ratios. The quantitative results are reported in Table~\ref{tab:lpips_packet_loss}. As the packet loss ratio increases from 0\% to 50\%, LPIPS rises gradually from 0.1748 to 0.3296. The degradation remains smooth across the tested range, without showing any abrupt quality collapse as packet loss becomes more severe. This indicates that our method preserves relatively stable visual reconstruction quality even under high packet loss, demonstrating meaningful tolerance to lossy transmission.

This behavior suggests that GenTrans exhibits graceful degradation when the received input is incomplete. Even when part of the transmitted information is lost, the system can still rely on the remaining conditions and learned generative priors to maintain perceptually usable reconstruction quality, rather than depending too strongly on fully received input. For real-world deployment, such smooth degradation is often more important than achieving the best result under ideal links, because network fluctuations are inevitable in practice and system usability largely depends on whether acceptable output can still be maintained under imperfect transmission.

Figure~\ref{fig:packet_loss_reconstruction} provides qualitative examples under different packet loss ratios. Even at 10\%, 30\%, and 50\% packet loss, the reconstructed frames still preserve the main scene structure and the dominant visual content. Although visual quality decreases as packet loss increases, the degradation remains progressive, without obvious failure cases or complete loss of recognizability. These visual observations are consistent with the LPIPS trends in Table~\ref{tab:lpips_packet_loss}, and further confirm the robustness of the proposed method under weak-network conditions.

\section{Conclusion}
In this paper, we proposed Generative Transmission (GenTrans), an AI Flow based generative video communication framework for ultra-low-bandwidth and weak-network scenarios. Different from conventional video transmission methods that mainly optimize pixel-level fidelity, GenTrans follows the AI Flow principle of transmitting compact intelligence representations rather than redundant raw signals, and formulates video communication as a joint optimization problem over bandwidth, computation, memory, and transmission robustness. By integrating generative reconstruction, cross-clip generative memory, runtime state reuse, and weak-network-aware transport design, GenTrans reduces redundant transmission and repeated decoding while maintaining perceptually coherent video reconstruction.
Experimental results demonstrate that GenTrans achieves clear advantages in three aspects: transmission efficiency, decoding efficiency, and robustness under packet loss. In particular, GenTrans significantly lowers bitrate compared with conventional codecs and prior generative compression methods while preserving competitive perceptual quality. Moreover, state reuse effectively reduces decoding latency, and the proposed robustness design enables graceful quality degradation under lossy network conditions. These results show that GenTrans is a practical and promising framework for video communication in challenging real-world environments.

\begin{credits}
\subsubsection{\ackname}
The authors would like to thank Wenkang Chen, Chunbo Hua, Wenyi Wang, Jingyu Xu and Fangqiu Yi for their valuable contributions to the algorithm implementation and engineering deployment. The contributors are listed in alphabetical order by surname.

% \subsubsection{\discintname}
% The authors have no competing interests to declare that are
% relevant to the content of this article.
% It is now necessary to declare any competing interests or to specifically
% state that the authors have no competing interests. Please place the
% statement with a bold run-in heading in small font size beneath the
% (optional) acknowledgments\footnote{If EquinOCS, our proceedings submission
% system, is used, then the disclaimer can be provided directly in the system.},
% for example: The authors have no competing interests to declare that are
% relevant to the content of this article. Or: Author A has received research
% grants from Company W. Author B has received a speaker honorarium from
% Company X and owns stock in Company Y. Author C is a member of committee Z.
\end{credits}
\bibliographystyle{splncs04}
\bibliography{waica}

@article{hevc,
  title={Overview of the High Efficiency Video Coding {(HEVC)} standard},
  author={Sullivan, Gary J and Ohm, Jens-Rainer and Han, Woo-Jin and Wiegand, Thomas},
  journal={IEEE Transactions on Circuits and Systems for Video Technology},
  volume={22},
  number={12},
  pages={1649--1668},
  year={2012},
  publisher={IEEE}
}

@article{aiflow,
  title={{AI Flow}: Perspectives, scenarios, and approaches},
  author={An, Hongjun and Hu, Wenhan and Huang, Sida and Huang, Siqi and Li, Ruanjun and Liang, Yuanzhi and Shao, Jiawei and Song, Yiliang and Wang, Zihan and Yuan, Cheng and others},
  journal={Vicinagearth},
  volume={3},
  number={1},
  pages={1},
  year={2026},
  publisher={Springer}
}

@article{vvc,
  title={Overview of the Versatile Video Coding {(VVC)} Standard and Its Applications},
  author={Bross, Benjamin and Wang, Ye-Kui and Ye, Yan and Liu, Shan and Chen, Jianle and Sullivan, Gary J and Ohm, Jens-Rainer},
  journal={IEEE Transactions on Circuits and Systems for Video Technology},
  volume={31},
  number={10},
  pages={3736--3764},
  year={2021},
  publisher={IEEE}
}

@article{gvc,
  title={{Generative Video Compression}: towards 0.01\% compression rate for video transmission},
  author={Chen, Xiangyu and Luo, Jixiang and Xu, Jingyu and Yi, Fangqiu and Zhang, Chi and Li, Xuelong},
  journal={Vicinagearth},
  volume={3},
  number={1},
  pages={7},
  year={2026},
  publisher={Springer}
}

@article{h264,
  title={Overview of the {H.264/AVC} video coding standard},
  author={Wiegand, Thomas and Sullivan, Gary J and Bjontegaard, Gisle and Luthra, Ajay},
  journal={IEEE Transactions on circuits and systems for video technology},
  volume={13},
  number={7},
  pages={560--576},
  year={2003},
  publisher={IEEE}
}

@INPROCEEDINGS{mcl_jcv,
  author={Wang, Haiqiang and Gan, Weihao and Hu, Sudeng and Lin, Joe Yuchieh and Jin, Lina and Song, Longguang and Wang, Ping and Katsavounidis, Ioannis and Aaron, Anne and Kuo, C.-C. Jay},
  booktitle={2016 IEEE International Conference on Image Processing}, 
  title={{MCL-JCV}: A {JND}-based {H.264/AVC} {V}ideo {Q}uality {A}ssessment {D}ataset}, 
  year={2016},
  pages={1509-1513},
}

@inproceedings{lpips,
  title={The unreasonable effectiveness of deep features as a perceptual metric},
  author={Zhang, Richard and Isola, Phillip and Efros, Alexei A and Shechtman, Eli and Wang, Oliver},
  booktitle={Proceedings of the IEEE conference on computer vision and pattern recognition},
  pages={586--595},
  year={2018}
}

@article{aiflow_edge,
  title={{AI Flow} at the network edge},
  author={Shao, Jiawei and Li, Xuelong},
  journal={IEEE Network},
  year={2025},
  publisher={IEEE}
}

@article{CBM_tradeoff,
  title={Computation-Bandwidth-Memory Trade-offs: A Unified Paradigm for AI Infrastructure},
  author={Fan, Yuankai and Weng, Qizhen and Li, Xuelong},
  journal={arXiv preprint arXiv:2601.11577},
  year={2025}
}

@article{yuan2025information,
  title={Information Capacity: Evaluating the Efficiency of Large Language Models via Text Compression},
  author={Yuan, Cheng and Shao, Jiawei and Li, Xuelong},
  journal={arXiv preprint arXiv:2511.08066},
  year={2025}
}

@article{li2024measuring,
  title={Measuring the information of images},
  author={Li, Xuelong and He, Rubin},
  journal={Scientia Sinica Informationis},
  volume={54},
  number={6},
  pages={1558},
  year={2024},
  publisher={Science China Press}
}

@article{positive_incentive_noise,
  title={Positive-incentive noise},
  author={Li, Xuelong},
  journal={IEEE Transactions on Neural Networks and Learning Systems},
  volume={35},
  number={6},
  pages={8708--8714},
  year={2022},
  publisher={IEEE}
}

@misc{sora,
title={Video Generation Models as World Simulators},
year = {2024},
author={OpenAI},
url = {https://openai.com/index/video-generation-models-as-world-simulators/}
}

\end{document}